%% file: main.tex
\documentclass[runningheads]{llncs}
\usepackage[T1]{fontenc}
\usepackage{hyperref}
\usepackage{multirow}
\usepackage{xcolor}
\definecolor{eccvblue}{rgb}{0.12,0.49,0.85}
\hypersetup{colorlinks=true,citecolor=eccvblue}
\usepackage{graphicx}
\usepackage{amssymb}
\usepackage{amsmath}

\begin{document}
\title{SAM3-UNet: Simplified Adaptation of Segment Anything Model 3}
\titlerunning{SAM3-UNet}
\author{Xinyu Xiong\inst{1*} \and
Zihuang Wu\inst{2*} \and Lei Lu\inst{3,4*} 
\and Yufa Xia\inst{5,6}
}

\authorrunning{X. Xiong et al.}
\institute{$^1$Sun Yat-sen University
$^2$Jiangxi Normal University 
$^3$Hainan University \\
$^4$Dalian University of Technology
$^5$Chizhou University 
$^6$Wuhan University \\
}

\renewcommand{\thefootnote}{}
\footnotetext{\inst{*} Authors contributed equally to this work.}
\maketitle 

\begin{abstract}
In this paper, we introduce SAM3-UNet, a simplified variant of Segment Anything Model 3 (SAM3), designed to adapt SAM3 for downstream tasks at a low cost. Our SAM3-UNet consists of three components: a SAM3 image encoder, a simple adapter for parameter-efficient fine-tuning, and a lightweight U-Net-style decoder. Preliminary experiments on multiple tasks, such as mirror detection and salient object detection, demonstrate that the proposed SAM3-UNet outperforms the prior SAM2-UNet and other state-of-the-art methods, while requiring less than 6 GB of GPU memory during training with a batch size of 12. The code is publicly available at \url{https://github.com/WZH0120/SAM3-UNet}.
\end{abstract}

\section{Introduction}
Foundation segmentation models, such as the Segment Anything Model series~\cite{SAM1,SAM2,SAM3}, have achieved remarkable success in recent years. Their large parameter scales and high-quality pre-training datasets endow them with strong out-of-domain generalization capabilities. Compared with SAM1~\cite{SAM1} and SAM2~\cite{SAM2}, the recently proposed SAM3~\cite{SAM3} introduces support for concept (text) prompts and offers further advancements in interactive segmentation, providing the potential for more precise segmentation in a wide range of downstream tasks.

Although the text-prompting capability of SAM3 enables the segmentation of specific semantic categories without human involvement in real-time, it still exhibits several limitations.
First, when processing objects with fine-grained structures~\cite{DIS5K}, SAM3 often provides only coarse boundary predictions, achieving approximate localization rather than precise delineation. Second, for segmentation tasks involving context-dependent targets~\cite{CVPR17_DUTS}, text prompts may fail altogether. These limitations indicate that fine-tuning SAM3 to fully exploit its powerful pre-trained representations for downstream segmentation tasks remains an important and valuable research direction.

In this paper, we propose SAM3-UNet, a simplified variant of SAM3 that enables efficient fine-tuning for downstream segmentation tasks. The advantages of SAM3-UNet are summarized as follows:

\begin{itemize}
\item \textbf{Compact.} We retain only the image encoder of SAM3 and remove the remaining components, yielding a more concise and easily extensible framework while preserving strong performance.

\item \textbf{Efficient.} By relaxing the resolution constraints and adopting a lightweight decoder, our framework lowers the hardware requirements for fine-tuning.

\item \textbf{Effective.} Experiments on two benchmarks demonstrate that our framework outperforms existing methods.
\end{itemize}


\section{Method}
\begin{figure*}[t]
    \centering
    \includegraphics[width=1.0\linewidth]{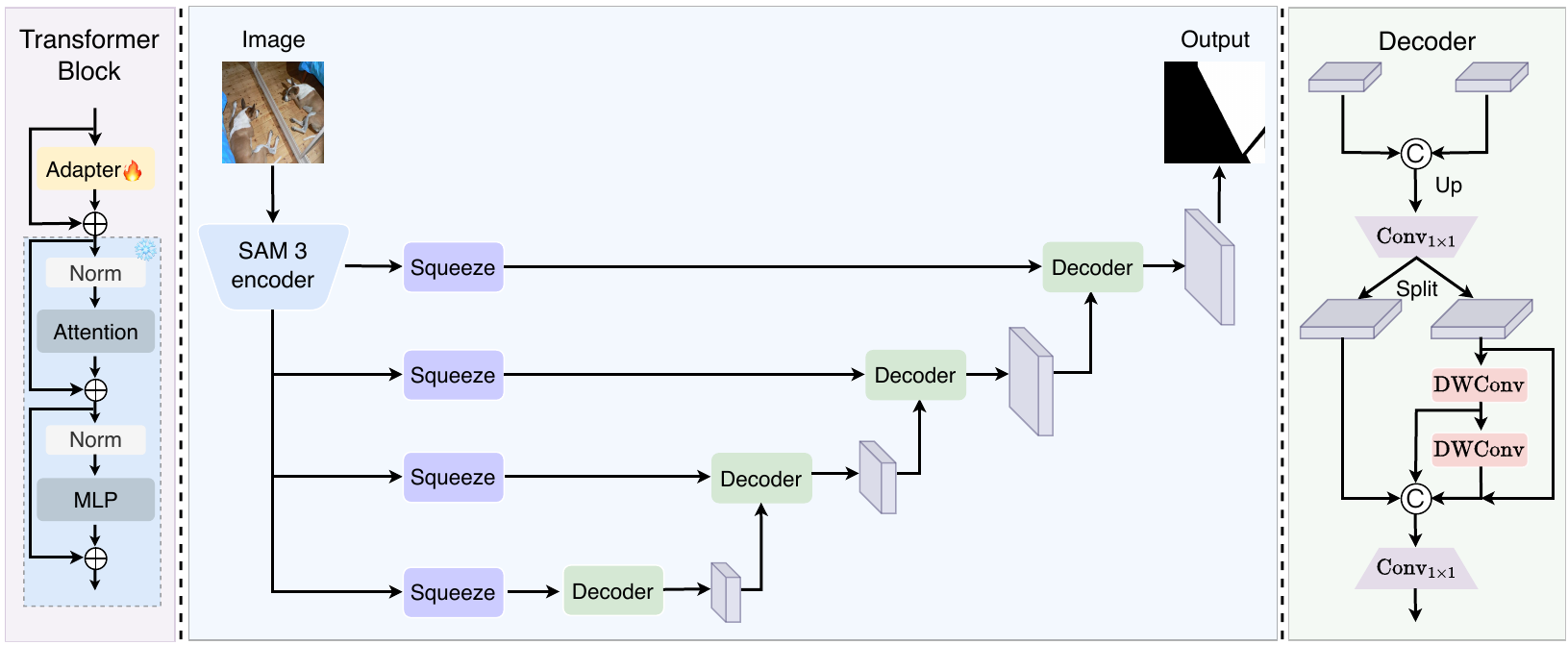}
    \caption{Overview of the proposed SAM3-UNet. For simplicity, we show only the decoder block where feature fusion is available.} 
    \label{fig:sam3unet}
\end{figure*}

The overall architecture of SAM3-UNet is shown in Fig.~\ref{fig:sam3unet}. It consists of an adapter-enhanced SAM3 image encoder and a lightweight U-Net–style decoder.

\subsection{SAM3 Encoder}
SAM3 employs a ViT-style~\cite{ICLR21_ViT} perception encoder~\cite{PE} as its image encoder, rather than continuing the Hiera~\cite{Hiera} architecture used in SAM2. In terms of scale, the vision encoder of SAM3 lies between the standard ViT-L and ViT-H, featuring a total of 446M parameters, which makes full fine-tuning computationally expensive. To address this, we design a simple yet effective parameter-efficient fine-tuning strategy. Specifically, we freeze all original ViT parameters and insert learnable adapter~\cite{Adapter} modules before each transformer block. Each adapter adopts a bottleneck structure composed of a downsampling layer, a GELU function, an upsampling layer, and a final GELU function.

In our previous work, SAM2-UNet~\cite{SAM2UNet}, the hierarchical encoder features are first enhanced using receptive-field blocks and then decoded by a standard U-Net~\cite{UNet} decoder composed of dual convolutional blocks. In this work, considering that the SAM3 encoder is non-hierarchical and larger than that of SAM2, we make the following modifications. First, for the ViT output embedding of size (H/14, W/14) with 1024 channels, where H and W denote the input image height and width, respectively, we apply four 1×1 convolutions to compress it into four lightweight feature maps, each with 128 channels. We then employ bilinear interpolation to resize these feature maps to (H/4, W/4), (H/8, W/8), (H/16, W/16), and (H/32, W/32), respectively. These resized features are treated as hierarchical representations and subsequently fed into a lightweight decoder for further learning.

\subsection{Lightweight Decoder}
In SAM2-UNet, the decoder adopts a standard double-convolution block design. Although this structure has been widely validated for its reliability and segmentation accuracy, its computational efficiency and convergence speed are inferior to those of more modern architectures. Therefore, while striving to maintain simplicity, we redesign a lightweight block to replace the original double-convolution block in the decoder.

\begin{figure*}[t]
    \centering
    \includegraphics[width=0.98\linewidth]{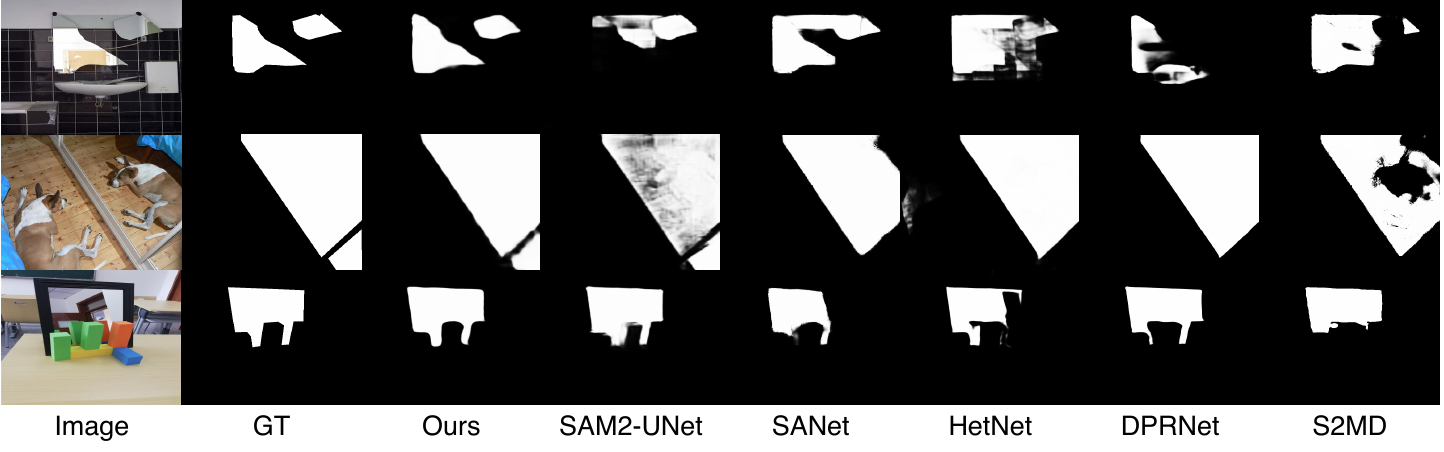}
    \caption{Visualization results on mirror detection.} 
    \label{fig:mirror}
\end{figure*}

\begin{figure*}[t]
    \centering
    \includegraphics[width=0.98\linewidth]{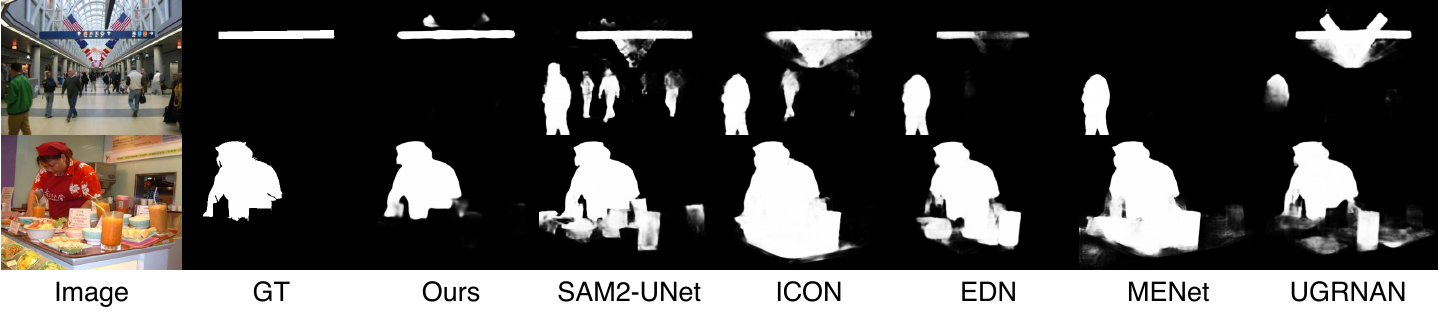}
    \caption{Visualization results on salient object detection.} 
    \label{fig:sod}
\end{figure*}

Specifically, to develop a simple yet efficient lightweight block, we incorporate three key design principles: bottlenecking, depthwise separable convolution, and feature splitting. Given an input feature map, we first apply a set of 1×1 Conv–BN–GELU layers to reduce the channel dimension from C to C/4. The reduced feature is then split evenly along the channel dimension into two parts, each with C/8 channels. The second part is sequentially processed by two 3×3 depthwise convolution blocks, generating the third and fourth parts. Finally, all four parts are concatenated along the channel dimension and passed through another set of 1×1 Conv-BN-GELU layers to expand the channels to the desired output dimension.

\section{Experiments}
\subsection{Datasets}

\input{tables/tab_mirror}

We conduct experiments on the following two segmentation benchmarks:

\textbf{Mirror Detection.} This benchmark conduct two sperate datasets. Following the official dataset split, for the MSD~\cite{ICCV19_MirrorNet} dataset, 3,063 images are used for training and 955 images are used for testing. For the PMD~\cite{CVPR20_PMD} dataset, 5,096 images are used for training and 571 are used for testing. We report results using IoU, F-measure (F), and mean absolute error (MAE).

\textbf{Salient Object Detection.} This benchmark consists five datasets in total. Following the common practice, the training set is DUTS-TR~\cite{CVPR17_DUTS}, which contains 10,553 images. The testing set contains five subsets: DUTS-TE~\cite{CVPR17_DUTS} (5,019 images), DUT-OMRON~\cite{CVPR13_DUTO} (5,168 images), HKU-IS~\cite{CVPR15_HKUIS} (4,447 images), PASCAL-S~\cite{CVPR14_PASCALS} (850 images), ECSSD~\cite{CVPR13_ECSSD} (1,000 images). We report results using S-measure~\cite{CVPR17_Smeasure} ($S_\alpha$), mean E-measure~\cite{Emeasure} ($E_{\phi}$), and mean absolute error (MAE).

\input{tables/tab_saliency}

\subsection{Implementation Details}
Our method is implemented in PyTorch and trained on a NVIDIA RTX 4090 GPU with 24 GB of memory. We use the AdamW optimizer with an initial learning rate of 0.0002 and apply cosine learning rate decay to stabilize training. The overall loss function consists of a weighted cross-entropy loss~\cite{AAAI20_F3Net} ($\mathcal{L}^{\omega}_{\text{BCE}}$) and a weighted IoU loss~\cite{AAAI20_F3Net} ($\mathcal{L}^{\omega}_{\text{IoU}}$). Two data augmentation strategies, including random horizontal and vertical flipping, are employed during training. The input resolutions are set to (H, W) = (336, 336). The bottleneck of the adapter module is set to 32 channels. All models are trained with a batch size of 12 for 20 epochs across different tasks.

\subsection{Result Analysis}
\subsubsection{Results on Mirror Detection.} As reported in Table~\ref{tab:mirror}, the proposed SAM3-UNet establishes a new state-of-the-art on both MSD and PMD benchmarks. It improves the best IoU from 0.918 to 0.943 on MSD (2.5\% improvement) and from 0.728 to 0.804 on PMD (7.6\% improvement). Some visual results are shown in Fig.~\ref{fig:mirror}. Our method is better at handling irregular shapes (row 1, 3) and hidden targets (row 2).

\subsubsection{Results on Salient Object Detection.} As reported in Table~\ref{tab:saliency}, taking the S-measure as an example, the proposed SAM3-UNet achieves better or comparable results on four out of the five datasets. The most significant improvement is observed on the DUT-OMRON dataset, where it further surpasses the second-best method, SAM2-UNet, by 1.1\%. Some visual results are shown in Fig.~\ref{fig:sod}. Our method more accurately identifies salient objects in cluttered environments (row 1, 2).

\section{Conclusion}
In this paper, we propose SAM3-UNet, a simple yet effective U-shaped framework that enables versatile segmentation through a simplified adaptation of Segment Anything Model 3. By retaining only the SAM3 image encoder, introducing lightweight adapters for parameter-efficient fine-tuning, and designing an efficient U-Net–style decoder, SAM3-UNet achieves strong segmentation performance with substantially reduced computational cost. Extensive experiments on multiple benchmarks demonstrate that our framework not only surpasses existing methods but also offers a practical and scalable solution for a wide range of downstream segmentation tasks.

\bibliographystyle{splncs04}
\bibliography{ref}
\end{document}

%% file: tables/tab_mirror.tex
\begin{table}[t]
\centering
\caption{Mirror detection performance on MSD~\cite{ICCV19_MirrorNet} and PMD~\cite{CVPR20_PMD} datasets.}
\label{tab:mirror}
\renewcommand\arraystretch{1.2}
\renewcommand\tabcolsep{2pt}
\begin{tabular}{l|ccc|ccc}
\hline
    & 
    \multicolumn{3}{c|}{MSD} & \multicolumn{3}{c}{PMD}
    \\
    \multirow{-2}{*}{Methods} & $IoU$ & $F$ & MAE & $IoU$ & $F$ & MAE\\
    \hline
    MirrorNet~\cite{ICCV19_MirrorNet} & 0.790 & 0.857 & 0.065 & 0.585 & 0.741 & 0.043 \\
    SANet~\cite{CVPR22_SAMirror} & 0.798 & 0.877& 0.054 & 0.668 & 0.795 & 0.032\\
    HetNet~\cite{AAAI23_HetNet} & 0.828 & 0.906 & 0.043 & 0.690 & 0.814 & 0.029\\
    DPRNet~\cite{TCSVT24_DPRNet} & 0.866 & / & 0.033 & 0.721 & / & 0.026 \\
    S2MD~\cite{TCSVT25_S2MD} & 0.871 & 0.936 & 0.032 & 0.698 & 0.846 & 0.024 \\
    SAM2-UNet~\cite{SAM2UNet} & 0.918 & 0.957 & 0.022 & 0.728 & 0.826 & 0.027 \\
    \hline
    \textbf{SAM3-UNet} & \textbf{0.943} & \textbf{0.972} & \textbf{0.014} & \textbf{0.804} & \textbf{0.884} & \textbf{0.017} \\
    \hline
\end{tabular}
\end{table}

%% file: tables/tab_saliency.tex
\begin{table}[t]
\centering
\caption{Salient object detection performance on DUTS-TE~\cite{CVPR17_DUTS}, DUT-OMRON~\cite{CVPR13_DUTO}, HKU-IS~\cite{CVPR15_HKUIS}, PASCAL-S~\cite{CVPR14_PASCALS} and ECSSD~\cite{CVPR13_ECSSD} datasets.}
\label{tab:saliency}
\renewcommand\arraystretch{1.2}
\renewcommand\tabcolsep{2pt}
\resizebox{1.0\columnwidth}{!}{
\begin{tabular}{l|ccc|ccc|ccc|ccc|ccc}
\hline
    & 
    \multicolumn{3}{c|}{DUTS-TE} & \multicolumn{3}{c|}{DUT-OMRON} & \multicolumn{3}{c|}{HKU-IS} & 
    \multicolumn{3}{c|}{PASCAL-S} & \multicolumn{3}{c}{ECSSD}\\
    \multirow{-2}{*}{Methods} & $S_\alpha$ & $E_{\phi}$ & MAE & $S_\alpha$ & $E_{\phi}$ & MAE & $S_\alpha$ & $E_{\phi}$ & MAE & $S_\alpha$ & $E_{\phi}$ & MAE & $S_\alpha$ & $E_{\phi}$ & MAE\\
    \hline
    U2Net~\cite{PR20_U2Net} & 0.874  & 0.884 & 0.044 & 0.847  & 0.872 & 0.054 & 0.916  & 0.948 & 0.031 & 0.844  & 0.850 & 0.074 & 0.928  & 0.925 & 0.033\\
    ICON~\cite{TPAMI22_ICON} & 0.889  & 0.914 & 0.037 & 0.845  & 0.879 & 0.057 & 0.920  & 0.959 & 0.029 & 0.861  & 0.893 & 0.064 & 0.929  & 0.954 & 0.032\\
    EDN~\cite{TIP22_EDN} & 0.892  & 0.925 & 0.035 & 0.850 & 0.877  & 0.049 & 0.924 & 0.955 & 0.026 & 0.865 & 0.902 & 0.062 & 0.927  & 0.951 & 0.032\\
    MENet~\cite{CVPR23_MENet} & 0.905  & 0.937 & 0.028 & 0.850  & 0.891 & 0.045 & 0.927 & 0.966 & 0.023 & 0.872  & 0.913 & 0.054  & 0.928 & 0.954 & 0.031\\
    UGRNAN~\cite{TIP25_UGRAN} & 0.931 & \textbf{0.964} & 0.022 & 0.878 & 0.911 & 0.045 & \textbf{0.945} & \textbf{0.978} & \textbf{0.019} & 0.892 & 0.934 & 0.046 & \textbf{0.950} & \textbf{0.975} & 0.020\\
    SAM2-UNet~\cite{SAM2UNet} & 0.934 & 0.959 & 0.020 & 0.884 & 0.912 & 0.039 & 0.941 & 0.971 & \textbf{0.019} & 0.894 & 0.931 & 0.043 & \textbf{0.950} & 0.970 & 0.020 \\
    \hline
    \textbf{SAM3-UNet} & \textbf{0.936} & \textbf{0.964} & \textbf{0.019} & \textbf{0.895} & \textbf{0.921} & \textbf{0.034} & 0.939 & 0.968 & 0.020 & \textbf{0.904} & \textbf{0.939} & \textbf{0.038} & \textbf{0.950} & 0.970 & \textbf{0.019}\\
    \hline
\end{tabular}
}
\end{table}